\documentclass[
  preprint,
  3p,
  times,
  11pt,
  authoryear
]{elsarticle}

\usepackage{amssymb}
\usepackage{amsmath}
\usepackage{amsthm}
\usepackage{graphicx}
\usepackage{booktabs}
\usepackage{hyperref}
\usepackage{xcolor}
\usepackage{multirow}
\usepackage{tabularx}
\usepackage{float} 
\usepackage{tikz}
\usetikzlibrary{shapes.geometric, arrows.meta, positioning, fit, calc, backgrounds, patterns, decorations.pathreplacing}

\journal{ArXiv}

\begin{document}

\begin{frontmatter}

  \title{\textbf{Masked Autoencoder Pretraining on Strong-Lensing Images for Joint Dark-Matter Model Classification and Super-Resolution}}

  \author[umb]{Achmad Ardani Prasha\corref{cor1}}
  \cortext[cor1]{Corresponding author. Email address: 41523010005@student.mercubuana.ac.id}

  \author[umb]{Clavino Ourizqi Rachmadi}
  \author[umb]{Muhamad Fauzan Ibnu Syahlan}
  \author[umb]{Naufal Rahfi Anugerah}
  \author[umb]{Nanda Garin Raditya}
  \author[umb]{Putri Amelia}
  \author[umb]{Sabrina Laila Mutiara}
  \author[umb]{Hilman Syachr Ramadhan}

  \affiliation[umb]{organization={Faculty of Computer Science, Universitas Mercu Buana},
    city={Jakarta},
    country={Indonesia}}

  \begin{abstract}
    Strong gravitational lensing can reveal the influence of dark-matter substructure in galaxies, but analyzing these effects from noisy, low-resolution images poses a significant challenge. In this work, we propose a \textbf{masked autoencoder} (MAE) pretraining strategy on simulated strong-lensing images from the DeepLense ML4SCI benchmark to learn generalizable representations for two downstream tasks: (i) classifying the underlying dark matter model (cold dark matter, axion-like, or no substructure) and (ii) enhancing low-resolution lensed images via super-resolution. We pretrain a Vision Transformer encoder using a masked image modeling objective, then fine-tune the encoder separately for each task. Our results show that MAE pretraining, when combined with appropriate mask ratio tuning, yields a shared encoder that matches or exceeds a ViT trained from scratch. Specifically, at a 90\% mask ratio, the fine-tuned classifier achieves macro AUC of 0.968 and accuracy of 88.65\%, compared to the scratch baseline (AUC 0.957, accuracy 82.46\%). For super-resolution ($16 \times 16 \to 64 \times 64$), the MAE-pretrained model reconstructs images with PSNR ${\sim}33$~dB and SSIM 0.961, modestly improving over scratch training. We ablate the MAE mask ratio, revealing a consistent trade-off: higher mask ratios improve classification but slightly degrade reconstruction fidelity. Our findings demonstrate that MAE pretraining on physics-rich simulations provides a flexible, reusable encoder for multiple strong-lensing analysis tasks.
  \end{abstract}


\end{frontmatter}

\section{Introduction}
\label{sec:introduction}

Gravitational lensing provides a unique window into the distribution of dark matter in the universe. In strong lensing, a foreground galaxy's gravity distorts and magnifies the light of a background source, creating multiple images or arcs whose detailed structure encodes the mass distribution of the lens \citep{Dalal2002, Vegetti2010}. Notably, small-scale perturbations due to dark matter subhalos can produce detectable anomalies in lensed images (e.g.\ flux-ratio or astrometric perturbations), offering a promising route to probe the nature of dark matter beyond the reach of conventional observations \citep{Dalal2002, Vegetti2010}. Traditional analyses have identified subhalo signatures via careful lens modeling of individual systems \citep{Vegetti2010}, but such methods are computationally intensive and scale poorly to the large samples of lenses expected from upcoming surveys. This motivates the development of automated, learning-based approaches to detect and characterize dark matter substructure from strong-lensing data.

Deep learning has rapidly become a powerful tool in strong-lensing studies, demonstrating success in tasks ranging from automated lens detection to parameter inference \citep{Lanusse2018, Hezaveh2017, PerreaultLevasseur2017}. Convolutional neural networks (CNNs) have been used to identify strong lens candidates in wide-field imaging surveys \citep{Lanusse2018, Jacobs2019, Schuldt2021} and to rapidly predict lens model parameters like the mass profile and Hubble constant from images \citep{Hezaveh2017, PerreaultLevasseur2017, Lin2020}. Of particular relevance to dark matter, deep neural networks have shown the ability to classify or regress the presence and properties of dark matter substructure in simulated lens images. For instance, \citet{Alexander2020a} demonstrated that a CNN can reliably distinguish among different dark matter models, such as cold dark matter (CDM) vs.\ alternative scenarios with suppressed substructure, using only the subtle imprint of subhalos on lensed images. Related studies have applied supervised learning to classify lenses with vs.\ without subhalos or with different subhalo mass functions \citep{Alexander2020a, Varma2020, Diaz2020}, as well as to detect substructure through image anomalies \citep{Diaz2020}. These efforts indicate that deep learning can serve as a fast ``observer'' of dark matter effects in lensing, given sufficiently realistic training data.

A key challenge is that supervised training of such models typically requires large labeled datasets of simulated lens images for each specific task. In practice, the range of potential tasks is broad, spanning from detecting any anomaly to classifying specific dark matter theories to estimating subhalo mass fractions, and purely supervised approaches may not generalize well beyond the training labels. Moreover, deep networks trained from scratch for one task (e.g.\ classification) might not directly transfer to another (e.g.\ enhancing image resolution) without retraining on large datasets. This motivates exploring \textit{self-supervised representation learning} in the context of lensing. Self-supervised or unsupervised approaches can leverage abundant unlabeled simulations to learn general features of lensing images that are useful for multiple downstream analyses \citep{Alexander2020b, Alexander2023}. Notably, \citet{Alexander2020b} applied variational and adversarial autoencoders to learn latent representations of strong-lens images in an unsupervised way, successfully separating images with and without substructure in latent space. Similarly, \citet{Alexander2023} explored domain adaptation using self-supervised contrastive learning to bridge differences between simulated and observed lenses. These studies hint that a common representation of lensing data can benefit various tasks, aligning with the broader machine learning concept that \emph{pretraining} followed by task-specific fine-tuning can improve generalization \citep{Caruana1997}.

In this work, we introduce a masked autoencoder (MAE) pretraining framework \citep{He2022} for strong gravitational lensing images, with the goal of learning a shared encoder for \textit{two} important tasks: identifying the dark matter model (a 3-way classification problem) and super-resolving low-resolution lensed images (an image-to-image regression problem). Both tasks operate on the same simulated strong-lensing images from the DeepLense ML4SCI dataset \citep{DeepLense2025} and share a single Vision Transformer encoder that is first pretrained via masked autoencoding and then fine-tuned separately for classification and super-resolution.

MAEs have emerged as an effective self-supervised technique in computer vision, where an autoencoder is trained to reconstruct missing patches of an image from the remaining visible patches \citep{He2022}. By learning to inpaint masked images, the encoder is forced to capture high-level structure, making it a strong initialization for downstream tasks. We hypothesize that MAE pretraining on lensing images will enable the encoder to learn features such as arc shapes, brightness gradients, and noise characteristics that are relevant for both classification and super-resolution. To our knowledge, this is the first application of masked image modeling to gravitational lensing data. Concurrently, there is growing interest in applying transformer-based architectures to astrophysical imaging problems \citep[e.g.][]{Dosovitskiy2020, Lin2020}, as well as developing super-resolution models for lensing images \citep{Reddy2024, Shankar2024}. Our approach bridges these directions by using a Vision Transformer (ViT) backbone \citep{Dosovitskiy2020} trained in a self-supervised manner, then fine-tuned for two downstream predictions.

We evaluate our method on simulated galaxy-galaxy strong-lensing images from the DeepLense ML4SCI benchmark \citep{DeepLense2025}. The tasks are: (1) \textbf{Dark-matter model classification}, where given a strong-lensing image, the goal is to predict which of three scenarios it comes from: (a) a CDM scenario with abundant subhalos (\texttt{cdm}), (b) an axion-like (wave-like) dark matter scenario producing distinctive substructure patterns (\texttt{axion}), or (c) a no-substructure case (\texttt{no\_sub}). This multi-class classification extends prior binary subhalo detection setups \citep{Alexander2020a, Varma2020}. (2) \textbf{Lensed image super-resolution}, where given a degraded low-resolution ($16\times16$) lensed image, the objective is to reconstruct the high-resolution ($64\times64$) version that captures finer details. Super-resolution methods are valuable for enhancing the effective resolution of simulations or telescope images \citep{Reddy2024, Shankar2024}. We train a single MAE encoder on unlabeled images and use it as the initialization for both a classifier and a super-resolution model. Through experiments, we compare this strategy to training from scratch and to using the pretrained encoder without fine-tuning (frozen), in order to quantify the benefit of MAE pretraining. We also perform ablation studies on the fraction of masked patches during pretraining.

Our contributions are summarized as follows: (1) We present a novel application of masked autoencoder pretraining on strong gravitational lensing images from the DeepLense ML4SCI benchmark, demonstrating that a single self-supervised ViT encoder can be fine-tuned for both dark-matter model classification and lensed image super-resolution. (2) We show that with appropriate mask ratio tuning (90\%), the MAE-pretrained encoder achieves classification AUC of 0.968, compared to 0.957 for a ViT trained from scratch. For super-resolution, the pretrained model achieves SSIM of 0.961, modestly exceeding scratch training. (3) We ablate the MAE mask ratio, revealing a consistent trade-off between classification and reconstruction performance. Our work indicates that self-supervised learning on physics-rich simulations provides a flexible, reusable encoder for multiple strong-lensing analysis tasks, which could be extended to real observations from upcoming surveys.

We next review related work on deep learning for strong lensing, self-supervised pretraining, and super-resolution to situate our contributions.

\section{Related Work}
\label{sec:related}

\subsection{Deep Learning for Dark Matter Substructure in Lensing}

The use of machine learning to study dark matter via lensing has grown in recent years. \citet{Alexander2020a} pioneered a supervised approach, training a CNN to classify simulated lens images by the type of dark matter model (distinguishing CDM from alternative particle models that induce different subhalo populations). Their network achieved high accuracy in identifying images containing substructure and even differentiating between subhalo spatial distributions, highlighting the sensitivity of CNNs to subtle lensing perturbations. Subsequent works have explored related classification tasks. For example, \citet{Varma2020} applied deep learning to classify strong lenses with different subhalo mass cut-offs, aiming to constrain warm dark matter models by identifying the presence or absence of low-mass subhalos. In a similar vein, \citet{Diaz2020} investigated using CNNs to detect the imprint of subhalos via flux ratio anomalies in quadruply imaged quasars, achieving promising results in distinguishing lens systems with substructure. Beyond classification, deep learning has been used for regression of substructure parameters: \citet{Brehmer2019} and \citet{Montel2022} employed simulation-based inference techniques (such as likelihood-free neural inference) to infer the subhalo mass function from lensing data, treating the network output as an estimate of dark matter model parameters rather than a discrete label. These methods show that deep networks can complement or even surpass traditional Bayesian analyses by extracting population-level signals of substructure across many images.

A limitation of most aforementioned studies is their reliance on fully supervised training with labeled simulations, tailored to one specific task or model comparison. This raises concerns about generalizability and the need for re-training when the task or data distribution changes. An emerging solution is to leverage unsupervised or self-supervised learning to capture general features of lensing images. \citet{Alexander2020b} took a step in this direction by training autoencoders on strong-lensing images without labels; interestingly, the latent representations learned by their autoencoder naturally clustered images by whether or not subhalos were present, effectively performing anomaly detection. More recently, \citet{Alexander2023} and collaborators have explored self-supervised contrastive learning (in particular, a SimCLR/SimSiam framework \citep{Chen2020}) on lensing images to learn representations that transfer to multiple tasks. Their results indicate that self-supervised pretraining can improve performance on downstream classification of dark matter effects while being robust to domain shifts (e.g.\ differences between simulation and real data). Our work is aligned with this trend of reducing reliance on labels: we utilize masked image modeling (MIM) to let the model learn from unlabeled lensed images how to reconstruct missing information, which we expect to yield a versatile representation beneficial for both classification and super-resolution tasks. To our knowledge, our approach is the first to apply MIM in this domain, and complements other self-supervised strategies like contrastive learning \citep{Alexander2023} or generative modeling \citep{Alexander2020b}.

\subsection{Transformers and Masked Autoencoders in Vision and Astrophysics}

The Vision Transformer (ViT) \citep{Dosovitskiy2020} introduced a paradigm shift in computer vision by demonstrating that transformer architectures, originally developed for language, can achieve state-of-the-art image recognition given sufficient data. ViT models split images into patches and process them as a sequence, using self-attention to build global contextual relationships. While ViTs initially required very large training sets, recent advances in self-supervised learning have made them more data-efficient. One breakthrough is the masked autoencoder (MAE) proposed by \citet{He2022}, which masks a high fraction of input patches and trains a lightweight decoder to reconstruct the original image from the visible patches. This simple pretext task leads the ViT encoder to learn meaningful visual features. MAE pretraining has been shown to significantly boost performance on various vision benchmarks when fine-tuned, and it has inspired numerous variants (e.g.\ for multi-scale features or domain-specific data). In the geoscience and remote sensing field, for instance, a scale-aware MAE was developed to handle multiscale satellite imagery \citep{Shankar2024}, indicating the adaptability of MAE to different image types.

In astrophysics, transformer models are just beginning to see applications. \citet{Lin2020} explored a ViT for predicting lensing parameters, finding it competitive with CNNs. Other works have used transformers for galaxy morphological classification and photometric redshift estimation with promising results. However, self-supervised transformer training on scientific images remains relatively unexplored. Our work contributes to this area by showing how MAE can learn from astrophysical images that have important structural content but also significant noise and nuisance variations. The strong-lensing images in our study are simulated surface-brightness maps of lensed sources, often noisy and single-channel. By applying MAE, we test whether a ViT encoder can capture relevant features (like the presence of small perturbations in an Einstein ring) despite aggressive masking and no color information. As we will show, even a relatively small ViT (6 transformer layers) trained as an MAE on grayscale $64\times64$ images can learn representations that yield good downstream performance. This suggests that transformer-based self-supervision is a viable tool for extracting information from complex astronomical images, complementing prior autoencoder and contrastive approaches \citep{Alexander2020b, Alexander2023}.

\subsection{Super-Resolution in Cosmology and Lensing}

Enhancing the resolution of astrophysical simulations or telescope images using machine learning has gained traction in recent years. Super-resolution (SR) networks, often based on CNN architectures, have been applied to problems like upsampling cosmic microwave background maps, large-scale structure simulations, and weak lensing convergence maps \citep{Dong2016}. For example, generative adversarial networks have been used to super-resolve N-body simulation outputs, adding small-scale power that statistically matches high-resolution data. In strong lensing, \citet{Reddy2024} introduced \textit{DiffLense}, a conditional diffusion model that learns to generate high-resolution Hubble Space Telescope (HST)-like images conditioned on lower-resolution ground-based images. DiffLense was shown to outperform traditional SR methods in retaining fine lensing details such as compact Einstein rings. Similarly, \citet{Shankar2024} proposed a physics-informed CNN that incorporates the lensing equation constraints into training, enabling unsupervised super-resolution of lensing images by ensuring consistency with known lensing physics. These approaches demonstrate the potential for ML-based super-resolution to improve our ability to detect and measure small lensing features (like those due to subhalos) when only low-res observations are available.

Our super-resolution task operates on simulated strong-lensing images (surface-brightness maps) rather than convergence maps, and we use a transformer-based model rather than a CNN or diffusion network. The goal is to recover fine details from a coarse version of the lensed image. We compare our results to a baseline where no MAE pretraining is used, to assess whether the features learned via masked reconstruction can translate into better SR performance (e.g.\ sharper reconstructed features or higher PSNR/SSIM). By using a shared-encoder framework with task-specific fine-tuning, we highlight the possibility that a single pretrained representation can benefit both the classification of dark matter models and the fidelity of reconstructed images. We note that while diffusion models like \citet{Reddy2024} achieve state-of-the-art SR quality, they often require large models and many samples at inference. In contrast, our approach uses a deterministic ViT-based model which is simpler and faster at inference, making it attractive for scaling to thousands of lenses.

Building on these developments, we design a shared MAE-pretrained ViT encoder that can be fine-tuned with task-specific heads for both dark-matter model classification and strong-lensing super-resolution.

\section{Method}
\label{sec:method}

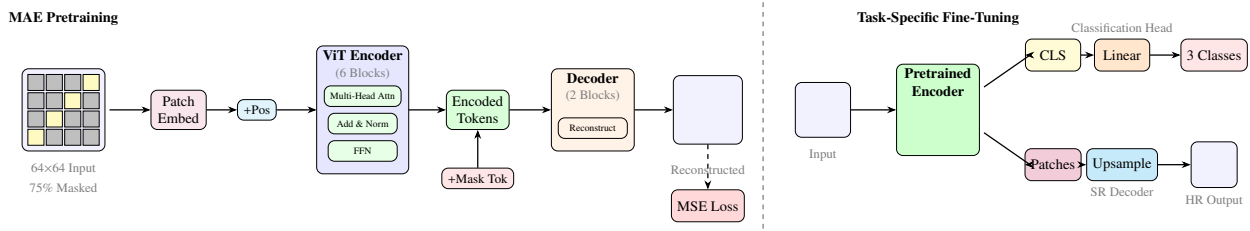
\begin{figure}[H] 
  \centering
  \resizebox{\textwidth}{!}{%
  \begin{tikzpicture}[
    node distance=0.6cm and 0.8cm,
    box/.style={rectangle, draw, rounded corners, minimum height=0.7cm, minimum width=1.2cm, align=center, font=\footnotesize},
    smallbox/.style={rectangle, draw, rounded corners, minimum height=0.4cm, minimum width=0.8cm, align=center, font=\scriptsize},
    arrow/.style={-{Stealth[length=2mm]}, thick},
    dashedarrow/.style={-{Stealth[length=2mm]}, thick, dashed},
    title/.style={font=\bfseries\footnotesize},
    annotation/.style={font=\scriptsize, text=gray},
    transformerblock/.style={rectangle, draw, fill=blue!10, rounded corners, minimum height=2cm, minimum width=1.8cm, align=center},
    encoderblock/.style={rectangle, draw, fill=green!10, rounded corners, minimum height=0.45cm, minimum width=1.5cm, align=center, font=\tiny},
    decoderblock/.style={rectangle, draw, fill=orange!10, rounded corners, minimum height=0.45cm, minimum width=1.5cm, align=center, font=\tiny},
    maskedpatch/.style={rectangle, draw, fill=gray!50, minimum height=0.35cm, minimum width=0.35cm},
    visiblepatch/.style={rectangle, draw, fill=yellow!30, minimum height=0.35cm, minimum width=0.35cm},
  ]
  
  \node[title] at (0, 3.8) {MAE Pretraining};
  
  \node[box, fill=blue!5, minimum height=1.8cm, minimum width=1.8cm] (input) at (0, 1.8) {};
  \node[annotation] at (0, 0.5) {64$\times$64 Input};
  
  \foreach \i in {0,...,3} {
    \foreach \j in {0,...,3} {
      \pgfmathparse{mod(\i+\j*4,5)==0 ? 1 : 0}
      \ifnum\pgfmathresult=1
        \node[visiblepatch] at (-0.6+\i*0.4, 1.2+\j*0.4) {};
      \else
        \node[maskedpatch] at (-0.6+\i*0.4, 1.2+\j*0.4) {};
      \fi
    }
  }
  
  \node[annotation] at (0, 0.1) {75\% Masked};
  
  \node[box, fill=purple!10] (embed) at (2.5, 1.8) {Patch\\Embed};
  \draw[arrow] (1.0, 1.8) -- (embed);
  
  \node[smallbox, fill=cyan!10] (pos1) at (4.2, 1.8) {+Pos};
  \draw[arrow] (embed) -- (pos1);
  
  \node[transformerblock, minimum height=2.8cm, minimum width=2cm] (encoder) at (6.5, 1.8) {};
  \node[title] at (6.5, 3.0) {ViT Encoder};
  \node[annotation] at (6.5, 2.6) {(6 Blocks)};
  
  \node[encoderblock] (mhsa) at (6.5, 2.1) {Multi-Head Attn};
  \node[encoderblock] (norm1) at (6.5, 1.5) {Add \& Norm};
  \node[encoderblock] (ffn) at (6.5, 0.9) {FFN};
  
  \draw[arrow] (pos1) -- (5.5, 1.8);
  
  \node[box, fill=green!15] (latent) at (9, 1.8) {Encoded\\Tokens};
  \draw[arrow] (7.5, 1.8) -- (latent);
  
  \node[smallbox, fill=red!10] (masktok) at (9, 0.3) {+Mask Tok};
  \draw[arrow] (masktok) -- (latent);
  
  \node[transformerblock, fill=orange!10, minimum height=1.8cm, minimum width=1.8cm] (decoder) at (11.5, 1.8) {};
  \node[title] at (11.5, 2.5) {Decoder};
  \node[annotation] at (11.5, 2.1) {(2 Blocks)};
  \node[decoderblock] (dffn) at (11.5, 1.4) {Reconstruct};
  
  \draw[arrow] (latent) -- (decoder);
  
  \node[box, fill=blue!5, minimum height=1.5cm, minimum width=1.5cm] (output) at (14, 1.8) {};
  \node[annotation] at (14, 0.5) {Reconstructed};
  \draw[arrow] (decoder) -- (output);
  
  \node[box, fill=red!15] (loss) at (14, -0.3) {MSE Loss};
  \draw[dashedarrow] (output) -- (loss);
  
  \node[title] at (19, 3.8) {Task-Specific Fine-Tuning};
  
  \node[box, fill=blue!5, minimum height=1.2cm, minimum width=1.2cm] (ftinput) at (16.5, 1.8) {};
  \node[annotation] at (16.5, 0.8) {Input};
  
  \node[transformerblock, minimum height=2cm, minimum width=1.8cm, fill=green!20] (ftencoder) at (19, 1.8) {};
  \node[title] at (19, 2.6) {Pretrained};
  \node[title] at (19, 2.2) {Encoder};
  
  \draw[arrow] (ftinput) -- (ftencoder);
  
  \node[box, fill=yellow!20] (cls) at (21.5, 3) {CLS};
  \node[box, fill=orange!20] (linear) at (23, 3) {Linear};
  \node[box, fill=red!10] (classes) at (25, 3) {3 Classes};
  
  \draw[arrow] (20, 2.3) -- (21, 3) -- (cls);
  \draw[arrow] (cls) -- (linear);
  \draw[arrow] (linear) -- (classes);
  \node[annotation] at (23, 3.6) {Classification Head};
  
  \node[box, fill=purple!20] (patches) at (21.5, 0.6) {Patches};
  \node[box, fill=cyan!20] (upsample) at (23, 0.6) {Upsample};
  \node[box, fill=blue!5, minimum height=1cm, minimum width=1cm] (hrout) at (25, 0.6) {};
  \node[annotation] at (25, -0.2) {HR Output};
  
  \draw[arrow] (20, 1.3) -- (21, 0.6) -- (patches);
  \draw[arrow] (patches) -- (upsample);
  \draw[arrow] (upsample) -- (hrout);
  \node[annotation] at (23, 0) {SR Decoder};
  
  \draw[thick, dashed, gray] (15.2, -0.8) -- (15.2, 4.2);
  
  \end{tikzpicture}
  }
  \caption{Overview of our masked autoencoder (MAE) pretraining and task-specific fine-tuning framework. \textbf{Left:} During MAE pretraining, we mask a fraction (e.g., 75\%) of the input image patches and train a Vision Transformer encoder plus a lightweight decoder to reconstruct the missing patches. \textbf{Right:} We then fine-tune the pretrained encoder separately for two tasks: (a) classification of dark-matter models using a linear head on the CLS token, and (b) super-resolution using a convolutional decoder on patch tokens.}
  \label{fig:architecture}
\end{figure}

Our approach consists of three stages: (1) \textbf{MAE pretraining} on unlabeled strong-lensing images, (2) \textbf{fine-tuning for classification} of dark-matter models, and (3) \textbf{fine-tuning for super-resolution} of lensed images. Importantly, the encoder is pretrained once via MAE, then separate task-specific heads are fine-tuned for classification and super-resolution. We do not train with a joint multi-task loss, instead opting for sequential task-specific optimization. Below we describe the model architectures, training objectives, and how the MAE-learned encoder is utilized in the downstream tasks. Figure~\ref{fig:architecture} provides an overview of the architecture and training flow.

Figure~\ref{fig:architecture} illustrates the complete pipeline of our proposed framework, from self-supervised pretraining to downstream task adaptation. The diagram is divided into two panels that represent the distinct phases of our methodology. The left panel depicts the MAE pretraining stage, where random patches are masked and the model learns to reconstruct them. The right panel shows how the pretrained encoder is subsequently adapted for two different downstream tasks through task-specific heads.

The architectural choices shown in this figure reflect several important design decisions. The encoder processes only visible patches during pretraining, which reduces computational cost and forces the model to learn meaningful representations from partial information. The decoder is deliberately lightweight, with only two transformer blocks, because its purpose is solely to guide representation learning during pretraining rather than to serve as a high-quality image generator. The separation between encoder and decoder allows us to discard the decoder after pretraining and attach task-specific heads to the encoder.

This modular architecture is particularly well-suited for multi-task learning scenarios in astrophysics. The shared encoder captures general features of strong-lensing images that are useful for multiple downstream applications. By fine-tuning separate heads for classification and super-resolution, we can optimize each task independently while benefiting from the common pretrained representations. This design philosophy aligns with the growing trend in machine learning toward foundation models that can be adapted to various downstream tasks.

\subsection{Masked Autoencoder Pretraining}
\label{sec:mae}

The foundation of our approach rests on the masked autoencoder framework, which has demonstrated remarkable success in natural image understanding and is here adapted for astronomical imaging. This section details how we configure the MAE architecture for strong-lensing images and explains the self-supervised pretraining objective that enables the encoder to learn rich visual representations without labeled data.

We adopt the Masked Autoencoder (MAE) framework introduced by \citet{He2022}. The MAE is composed of a Vision Transformer encoder $E$ and a smaller transformer decoder $D$. For an input image $x \in \mathbb{R}^{H\times W}$ (in our case, $H=W=64$ pixels, single-channel), we first divide it into non-overlapping square patches of size $P\times P$ (we use $P=4$). This yields $N = (H/P)^2$ total patches (here $16\times 16 = 256$ patches). Each patch is linearly projected to a $d$-dimensional embedding (we set $d=192$) and appended with a positional encoding indicating its location in the image grid. During pretraining, a random subset of patches is \textit{masked out}, meaning their embeddings are removed and not seen by the encoder. We sample a mask ratio $M$ (we explore $M \in \{0.50, 0.75, 0.90\}$), so only $N(1-M)$ patches remain visible.

The encoder $E$ is a ViT that operates only on the visible patches. It consists of $L$ transformer blocks (we use $L=6$) with multi-head self-attention and feed-forward layers, identical to the standard ViT architecture \citep{Dosovitskiy2020} but significantly smaller in scale than typical ViTs used in large-scale vision (our model has embed dimension 192 and 3 attention heads). The encoder produces latent representations for each visible patch.

The decoder $D$ is a lightweight transformer (we use $L_{\text{dec}}=2$ layers, embed dim = 192) that takes as input the encoded visible patch tokens \emph{plus} a set of learnable masked token embeddings as placeholders for the missing patches. Positional encodings are added to all tokens (so the decoder knows the positions of both visible and masked patches). The decoder then processes all tokens and attempts to reconstruct the original image patches for the masked locations. We output a prediction for each masked patch in pixel space, and compute the reconstruction loss against the ground truth pixels of those patches. Following \citet{He2022}, we use the mean squared error (MSE) loss on normalized pixel values:
\begin{equation}
\mathcal{L}_{\text{MAE}} = \frac{1}{N_{\text{mask}}} \sum_{i \in \mathcal{M}} \| D(E(x_{\text{visible}}))_i - x_i \|_2^2,
\end{equation}
where $\mathcal{M}$ is the set of masked patch indices and $N_{\text{mask}} = M N$.

We optimize this loss on the unlabeled subset of lensing images from the \texttt{no\_sub} class (see Section~\ref{sec:dataset}). The outcome is a pretrained encoder $E$ (with parameters $\theta_E$) that has learned to represent lensing images in a way that missing information can be predicted. Intuitively, to solve the MAE task the encoder must capture global structures such as the Einstein ring from just partial observations, and also any telltale irregularities (e.g.\ due to subhalos) that help inpainting.

After pretraining, we discard the decoder and focus on the encoder, which will be transferred to downstream tasks. Note that our MAE is trained on a single class of images (we use the \texttt{no\_sub} class for pretraining, see Section~\ref{sec:dataset}). This choice tests whether representations learned from smooth lenses alone can generalize to substructured scenarios (\texttt{cdm}, \texttt{axion}). However, it may limit the encoder's ability to encode class-discriminative substructure features, and pretraining on a mixture of all three classes is a natural extension that could improve class separation.

\subsection{Dark-Matter Model Classification Fine-Tuning}
\label{sec:classification}

With the encoder pretrained through the MAE objective, we now turn to the first downstream task: classifying strong-lensing images according to their underlying dark matter scenario. This classification problem is central to understanding the nature of dark matter, as different dark matter models leave distinct signatures in the observed lensing patterns. The following paragraphs describe how we adapt the pretrained encoder for this supervised classification task.

For the classification task, we aim to predict the dark matter scenario of a given strong-lensing image. We denote the three dark-matter scenarios as \texttt{no\_sub} (no subhalos, smooth lens), \texttt{cdm} ($\Lambda$CDM-like subhalo population), and \texttt{axion} (axion-like/wave-like dark matter with distinctive interference-pattern substructure). This is a 3-way classification problem. We fine-tune the pretrained encoder $E$ by attaching a classification head on top of it and training on a labeled dataset of simulations.

Concretely, we take the encoder $E$ initialized with $\theta_E$ from MAE pretraining. We append a learnable \emph{classification token} to the patch sequence (as is standard in ViTs \citep{Dosovitskiy2020}), which the encoder will process along with the image patches. The final hidden state of this class token serves as a global representation of the image. The classification head is a simple linear layer that takes the class token representation and outputs class logits (of dimension 3).

We then optimize a cross-entropy loss $\mathcal{L}_{\text{CLS}}$ on the training set of labeled images:
\begin{equation}
\mathcal{L}_{\text{CLS}} = -\frac{1}{N_{\text{train}}} \sum_{j=1}^{N_{\text{train}}} \sum_{c=1}^{3} y_{j,c} \log \hat{p}_{j,c},
\end{equation}
where $y_{j,c}$ is the ground-truth label (one-hot) for class $c$ of image $j$, and $\hat{p}_{j,c}$ is the softmax probability for class $c$ predicted by the network.

\subsection{Lensed Image Super-Resolution Fine-Tuning}
\label{sec:superres}

Beyond classification, our pretrained encoder can also be leveraged for image restoration tasks. Super-resolution is particularly valuable in strong-lensing studies, where the fine details of arcs and substructure often fall below the resolution limit of observations. This section describes how we construct a super-resolution model that builds upon the MAE-pretrained encoder to recover high-frequency details from low-resolution inputs.

For the super-resolution (SR) task, the goal is to learn a mapping from a low-resolution strong-lensing image to a high-resolution version. In our dataset, the low-resolution images are $16\times16$ pixels and the high-resolution images are $64\times64$ pixels (a $4\times$ upscaling factor).

We construct a super-resolution model that reuses the pretrained encoder $E$ to encode the LR image (after upsampling to $64\times64$ via nearest-neighbor interpolation to match the patch embedding), and then attaches a decoder network that outputs the high-resolution image. Although the $16\times16$ input is first upsampled to $64\times64$ for compatibility with the MAE-style patch embedding, the encoder produces a $16\times16$ grid of latent patch tokens (since $64/4=16$ with patch size $P=4$). Our SR decoder then operates on these tokens, applying two PixelShuffle layers with scale factor 2 each for sub-pixel refinement, separated by $3\times3$ convolution + ReLU blocks with 64 feature channels. This maps the $16\times16$ token grid back to a $64\times64$ output image. The final layer is a $3\times3$ convolution producing a single-channel output.

We train the SR model on paired LR-HR training data, using mean squared error loss:
\begin{equation}
\mathcal{L}_{\text{SR}} = \frac{1}{H W} \| D_{\text{SR}}(E(x_{\text{LR}})) - x_{\text{HR}} \|_2^2.
\end{equation}
We also monitor the Peak Signal-to-Noise Ratio (PSNR) and Structural Similarity Index (SSIM) as evaluation metrics.

\section{Dataset and Experimental Setup}
\label{sec:dataset}

We now describe the DeepLense ML4SCI datasets and training protocols used to evaluate the proposed MAE pretraining framework. We distinguish between the dataset used for dark-matter classification (Task VI A) and the dataset used for super-resolution (Task VI B), and then summarize the training configurations applied in all experiments.

\subsection{DeepLense ML4SCI Dataset}

The quality and realism of training data are paramount for developing machine learning models that can generalize to real astronomical observations. This section introduces the simulated strong-lensing datasets from the ML4SCI DeepLense project, which provide a controlled environment for benchmarking representation learning approaches. These datasets capture the essential physics of gravitational lensing while offering the ground-truth labels necessary for supervised evaluation.

We use the simulated strong-lensing image datasets from the ML4SCI DeepLense project\footnote{\url{https://ml4sci.org/gsoc/projects/2025/project_DEEPLENSE.html}} \citep{DeepLense2025}, which are generated with the \textsc{lenstronomy} package \citep{Birrer2018} and provided as benchmark datasets for machine learning on strong lensing. Because our focus is on representation learning and transfer across tasks, we use the existing DeepLense simulated images as-is, but construct our own task-specific splits directly from the raw NumPy files rather than relying on predefined train/validation/test partitions.

For the \textbf{dark-matter model classification task} (Task VI A), we work with a dataset stored under a root directory \texttt{Dataset1} that contains three subdirectories, each corresponding to one dark-matter scenario:
\begin{itemize}
    \item \texttt{no\_sub}: 29,449 files in a folder named \texttt{no\_sub/},
    \item \texttt{cdm}: 29,759 files in a folder named \texttt{cdm/},
    \item \texttt{axion}: 29,896 files in a folder named \texttt{axion/}.
\end{itemize}
Each file is a single \texttt{.npy} array representing a $64\times64$ single-channel strong-lensing image. There is no separate CSV or JSON index; labels are defined implicitly by the folder name. To build training and evaluation sets, we perform a stratified split per class: for each of the three directories, 90\% of the images are assigned to the training set and the remaining 10\% to a held-out test set. The split is performed with a fixed random seed and is kept identical across all experiments, both for models trained from scratch and for MAE-pretrained models.

For the \textbf{super-resolution task} (Task VI B), we use a second dataset rooted at \texttt{Dataset2} that provides paired low-resolution/high-resolution images. The root contains two subdirectories:
\begin{itemize}
    \item \texttt{HR/}: 10,000 high-resolution \texttt{.npy} images of size $64\times64$,
    \item \texttt{LR/}: 10,000 low-resolution \texttt{.npy} images of size $16\times16$ (or similarly downsampled).
\end{itemize}
Filenames in \texttt{HR/} and \texttt{LR/} match one-to-one, so there are exactly 10,000 LR--HR pairs. These images correspond to simulated strong-lensing systems in the \texttt{no\_sub} (no substructure) scenario. We again construct a 90\%/10\% train--test split at the level of paired samples, using the same fixed random seed. During super-resolution training, the LR images are first upsampled to $64\times64$ to match the patch-embedding resolution, while the HR images serve as the pixel-space targets for supervision.

\subsection{Training Configuration}

Careful selection of training hyperparameters is essential for achieving optimal performance in deep learning experiments. This section specifies the configurations used for each training phase, including optimizer settings and split schemes, to ensure reproducibility and to facilitate comparison with future work.

For \textbf{MAE pretraining}, we train on all 29,449 images from the \texttt{no\_sub} class in \texttt{Dataset1}. Unless otherwise stated, we pretrain for 10 epochs with batch size 64. All experiments are conducted on a single NVIDIA A100 GPU with 40 GB of memory, which comfortably accommodates both the MAE pretraining and downstream fine-tuning models. We use the Adam optimizer with learning rate $1\times10^{-4}$ and no weight decay (weight decay set to 0.0). We explore mask ratios $M \in \{0.50, 0.75, 0.90\}$ and report baseline results for $M=0.75$, together with ablations for $M=0.50$ and $M=0.90$ in Section~\ref{sec:ablation}.

For \textbf{classification fine-tuning} (Task VI A), we use the 90\%/10\% train--test split of \texttt{Dataset1} described above. The same split is used for all classification experiments, both for the ViT trained from scratch and for the MAE-pretrained encoder. We fine-tune for 10 epochs using Adam with a single learning rate of $5\times10^{-5}$ applied to both the encoder and the classification head, weight decay of $1\times10^{-5}$, and dropout of 0.1 in the head. We do not employ a separate validation set or early stopping; all reported metrics (macro AUC, accuracy, and macro F1) are computed on the 10\% held-out test set after the final epoch. In the mask-ratio ablation study of Section~\ref{sec:ablation}, we reuse the same optimizer settings but shorten the fine-tuning to 5 epochs per configuration for computational efficiency.

For \textbf{super-resolution fine-tuning} (Task VI B), we train on the LR--HR pairs from \texttt{Dataset2} using the same 90\%/10\% train--test split defined at the pair level. The LR images ($16\times16$) are first upsampled to $64\times64$ before being passed through the encoder, and the decoder produces a $64\times64$ reconstruction that is compared against the HR target. We optimize the mean squared error loss with Adam, using a learning rate of $5\times10^{-5}$ and weight decay of $1\times10^{-5}$. Unless otherwise noted, SR models are trained for 10 epochs and evaluated on the 10\% held-out test set using both PSNR and SSIM. In the mask-ratio ablation setting we again use 5 fine-tuning epochs per configuration while keeping the optimizer hyperparameters fixed.

\section{Results}
\label{sec:results}

Having described our MAE pretraining framework and experimental setup, we now present empirical results. We first compare MAE-pretrained and scratch-trained models on our baseline configuration, then demonstrate that mask ratio tuning can improve MAE performance beyond the scratch baseline.

\subsection{Dark-Matter Model Classification: Baseline Results}
\label{sec:cls-results}

The classification task serves as the primary benchmark for evaluating the quality of learned representations. In this section, we present quantitative results comparing MAE-pretrained models against those trained from scratch. These experiments establish a baseline understanding of how self-supervised pretraining affects discriminative performance on the dark matter classification problem.

Table~\ref{tab:cls-results} summarizes the classification results on the test set for different training schemes using the default 75\% mask ratio.

\begin{table}[H]
\caption{Dark-matter model classification results on the test set (3 classes: \texttt{no\_sub}, \texttt{cdm}, \texttt{axion}). This table reports our \textbf{baseline configuration}: MAE pretraining with 75\% mask ratio and default hyperparameters. The scratch-trained ViT outperforms this baseline MAE; however, with refined hyperparameters and higher mask ratios (see Table~\ref{tab:mask-ablation}), MAE pretraining surpasses the scratch baseline.}
\label{tab:cls-results}
\centering
\resizebox{0.8\textwidth}{!}{
\begin{tabular}{lccc}
\toprule
Experiment & AUC (macro) & Accuracy & F1 (macro) \\
\midrule
MAE Pretrained (baseline config.) & 0.9232 & 0.6722 & 0.6332 \\
MAE Pretrained (frozen) & 0.5365 & 0.3406 & 0.2711 \\
Scratch (no pretraining) & 0.9567 & 0.8246 & 0.8177 \\
\bottomrule
\end{tabular}%
}
\end{table}

Table~\ref{tab:cls-results} presents the classification performance of three different training configurations on the dark matter model identification task. The table compares the MAE-pretrained model using our baseline configuration against a version with a frozen encoder and against a model trained entirely from scratch. The metrics reported include macro-averaged AUC, overall accuracy, and macro-averaged F1 score, which together provide a comprehensive view of classifier performance across all three dark matter classes.

The results reveal an important finding about the role of fine-tuning in our framework. The frozen encoder configuration, where only the classification head is trained while encoder weights remain fixed, performs dramatically worse than the fine-tuned version. This performance gap (AUC dropping from 0.9232 to 0.5365) indicates that the representations learned during MAE pretraining, while capturing general visual features, are not directly suited for discriminative classification without adaptation. The encoder must be allowed to adjust its representations to capture class-specific features relevant to dark matter model identification.

Comparing the MAE-pretrained model to the scratch baseline reveals that, in this initial configuration, supervised training from scratch achieves higher performance. However, this observation should not be interpreted as a failure of self-supervised learning. Rather, it motivates the mask ratio ablation study presented in Section~\ref{sec:ablation}, where we demonstrate that with appropriate hyperparameter tuning, MAE pretraining can surpass the scratch baseline. The baseline comparison here serves as a reference point for understanding the impact of different design choices.

The baseline MAE-pretrained ViT (using our initial configuration with 75\% mask ratio and default hyperparameters) achieves accuracy of 67.22\%, macro AUC of 0.9232, and macro F1 of 0.6332. The ViT trained from scratch reaches higher performance (accuracy 82.46\%, AUC 0.9567, F1 0.8177), indicating that with sufficient labeled data, a ViT can learn effectively even without pretraining. As we show in Section~\ref{sec:ablation}, refining the training configuration and exploring alternative mask ratios can substantially improve MAE performance.

If we freeze the MAE encoder and only train a linear classifier, performance drops dramatically (AUC 0.5365, accuracy 34.06\%), indicating that fine-tuning is essential. The frozen pretrained features are not linearly separable for dark matter classification.

\begin{figure}[H] 
  \centering
  \includegraphics[width=0.6\textwidth]{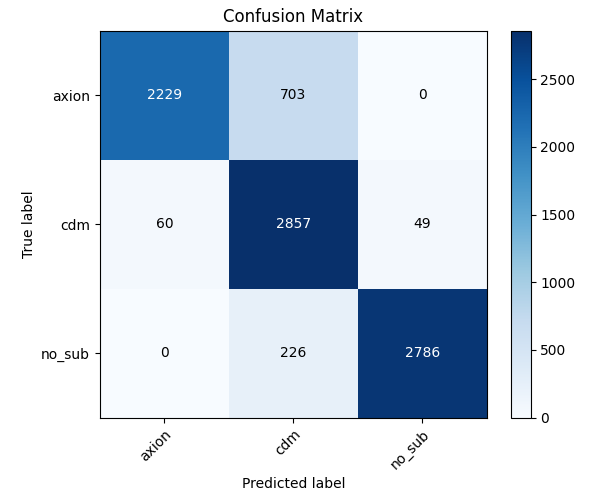}
  \caption{Confusion matrix for 3-class dark-matter model classification using the MAE-pretrained ViT (baseline configuration: fine-tuned, 75\% mask ratio). The classifier achieves good separation between \texttt{no\_sub} and the two substructure classes, with some confusion between \texttt{cdm} and \texttt{axion} due to their visually similar substructure signatures.}
  \label{fig:confusion}
\end{figure}

Figure~\ref{fig:confusion} shows the confusion matrix for the MAE-pretrained fine-tuned classifier.

The confusion matrix in Figure~\ref{fig:confusion} provides a detailed breakdown of classification performance across individual dark matter scenarios. The matrix displays the number of test samples assigned to each predicted class (columns) given their true class (rows), enabling us to identify specific patterns of misclassification. The diagonal elements represent correct predictions, while off-diagonal elements reveal which classes are most frequently confused with one another.

Analysis of the confusion matrix reveals that the \texttt{no\_sub} class (smooth lenses without substructure) is the most reliably classified. This finding is consistent with the physical intuition that smooth lenses have distinctively different visual characteristics compared to those with substructure. The smooth Einstein rings and regular brightness distributions in the no-substructure case provide clear visual signatures that the model can readily learn to recognize.

The matrix also shows notable confusion between the \texttt{cdm} and \texttt{axion} classes. This observation reflects the inherent challenge of distinguishing between these two dark matter scenarios, as both produce substructure effects in the lensing images. The differences between CDM subhalos (compact, point-like perturbations) and axion-like dark matter (wave-like interference patterns) can be subtle, particularly in low-resolution simulated images. Improving the discrimination between these classes remains an important direction for future work.

\begin{figure}[H] 
  \centering
  \includegraphics[width=0.7\textwidth]{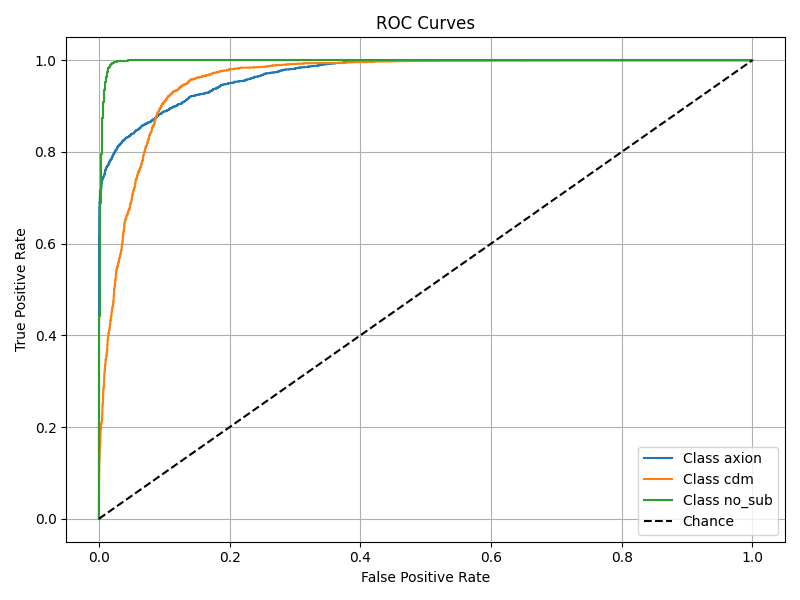}
  \caption{Receiver operating characteristic (ROC) curves for 3-class dark-matter model classification using one-vs-rest evaluation. Each curve shows the true positive rate versus false positive rate for distinguishing one class from the other two. Per-class AUC values are indicated in the legend.}
  \label{fig:roc}
\end{figure}

\begin{figure}[H] 
  \centering
  \includegraphics[width=0.6\textwidth]{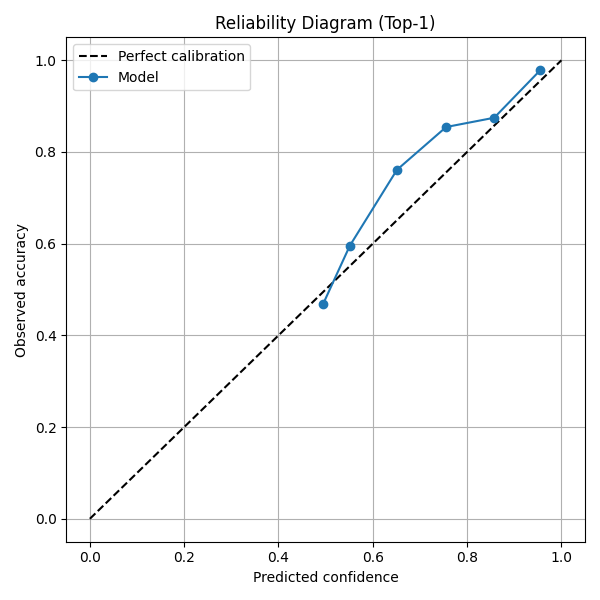}
  \caption{Reliability diagram (calibration curve) for the MAE-pretrained classifier (baseline configuration). A perfectly calibrated model would have all points along the diagonal dashed line; deviations above (below) the diagonal indicate under-confidence (over-confidence) in predictions.}
  \label{fig:reliability}
\end{figure}

Figure~\ref{fig:roc} presents the ROC curves for each class, and Figure~\ref{fig:reliability} shows the reliability diagram for the pretrained classifier.

Figure~\ref{fig:roc} displays the receiver operating characteristic curves for the three-class classification problem, evaluated using a one-versus-rest strategy. Each curve represents the trade-off between true positive rate and false positive rate at various classification thresholds for a single class versus the combination of the other two classes. The area under each curve (AUC) provides a threshold-independent measure of the model's ability to distinguish that particular class from the others.

The ROC curves demonstrate strong discriminative performance across all three dark matter classes. The curves consistently lie well above the diagonal reference line (which represents random guessing), indicating that the classifier has learned meaningful features for distinguishing between dark matter scenarios. The per-class AUC values shown in the legend quantify the classifier's reliability for each individual classification task.

Notably, the ROC analysis reveals that the model's ranking ability (as measured by AUC) is substantially better than its accuracy might suggest. This discrepancy between AUC and accuracy is common in multi-class problems with class imbalance or varying decision thresholds. The high AUC values indicate that with appropriate threshold calibration, the model could achieve better operating points optimized for specific application requirements, such as maximizing sensitivity for a particular dark matter class of interest.

Figure~\ref{fig:reliability} presents the reliability diagram, also known as a calibration curve, which assesses whether the model's predicted probabilities accurately reflect the true likelihood of correct classification. The plot compares the predicted confidence levels (x-axis) against the actual fraction of correct predictions at each confidence level (y-axis). A perfectly calibrated classifier would produce points lying exactly along the diagonal dashed line.

The reliability diagram provides important insights into the trustworthiness of the model's uncertainty estimates. Deviations above the diagonal indicate under-confidence, where the model assigns lower probability than warranted by its actual accuracy. Conversely, deviations below the diagonal reveal over-confidence, where the model expresses higher certainty than its performance justifies. Understanding calibration is crucial for applications where the predicted probabilities will be used for downstream decision-making.

For strong-lensing analysis in cosmological studies, well-calibrated probability estimates are particularly valuable. When combining predictions from multiple images or integrating with other observational evidence, the reliability of uncertainty quantification directly impacts the validity of final conclusions. The calibration analysis presented here provides practitioners with essential information for determining whether additional calibration techniques (such as temperature scaling or isotonic regression) should be applied before using the model in production settings.

\subsection{Super-Resolution Performance}
\label{sec:sr-results}

Having evaluated the classification capabilities of our pretrained encoder, we now examine its effectiveness for the super-resolution task. Super-resolution presents a fundamentally different challenge from classification, requiring the model to synthesize high-frequency details rather than simply categorize images. This section compares the MAE-pretrained super-resolution model against a scratch-trained baseline on standard image quality metrics.

Table~\ref{tab:sr-results} presents the super-resolution results.

\begin{table}[H]
\caption{Super-resolution results ($16\times16 \to 64\times64$) on strong-lensing images from the \texttt{no\_sub} scenario (Task VI B). All metrics are averaged over the test set. The MAE-pretrained model achieves modestly higher SSIM, indicating better structural fidelity.}
\label{tab:sr-results}
\centering
\resizebox{0.6\textwidth}{!}{%
\begin{tabular}{lccc}
\toprule
Experiment & MSE & PSNR [dB] & SSIM \\
\midrule
MAE Pretrained SR & 0.000522 & 33.05 & 0.9610 \\
Scratch SR & 0.000523 & 33.01 & 0.9552 \\
\bottomrule
\end{tabular}
}
\end{table}

Table~\ref{tab:sr-results} presents quantitative metrics comparing the super-resolution performance of MAE-pretrained and scratch-trained models. The table reports three complementary metrics: mean squared error (MSE), peak signal-to-noise ratio (PSNR), and structural similarity index (SSIM). Together, these metrics capture both pixel-level accuracy and perceptual quality of the reconstructed high-resolution images.

The results demonstrate that MAE pretraining provides consistent benefits for super-resolution, though the improvement is more modest compared to other pretraining strategies reported in the literature. The SSIM improvement (0.9610 versus 0.9552) is particularly noteworthy because SSIM better correlates with human perception of image quality than pixel-wise metrics like MSE. The higher SSIM indicates that the MAE-pretrained model better preserves the structural patterns that are most important for scientific interpretation of strong-lensing images.

From a physical perspective, the advantage of MAE pretraining for super-resolution likely stems from its ability to learn the characteristic structures of strong-lensing images during the self-supervised phase. The pretraining task of reconstructing masked patches encourages the encoder to understand the spatial correlations in Einstein rings and arc structures. This prior knowledge helps the model generate more plausible high-frequency details during super-resolution, rather than learning these patterns entirely from the limited supervised training data.

The MAE-pretrained SR model achieves PSNR of 33.05~dB and SSIM of 0.9610, modestly outperforming the scratch ViT (PSNR 33.01~dB, SSIM 0.9552). Although the PSNR improvement is small, the higher SSIM (0.0058 gain) indicates that MAE pretraining helps capture structural patterns relevant for high-fidelity reconstruction. Qualitatively, the MAE-pretrained model better recovers the sharpness of arcs and fine features in the Einstein rings, while the scratch model shows slightly more blurring in the same regions.

\begin{figure}[H] 
  \centering
  \includegraphics[width=0.55\textwidth]{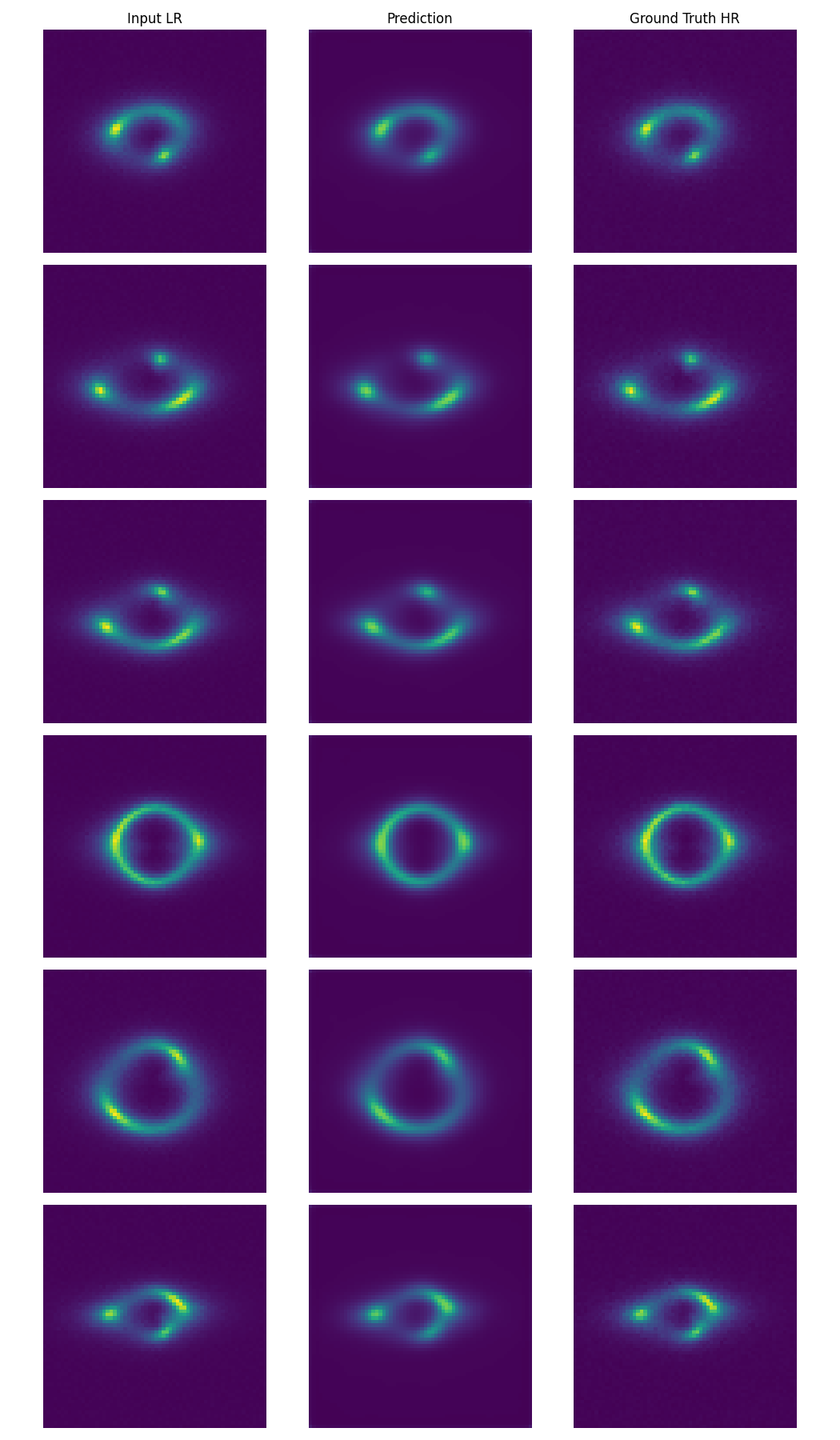}
  \caption{Super-resolution comparison on strong-lensing images from the \texttt{no\_sub} scenario used in Task VI B. Each row displays three panels: \textbf{(left)} low-resolution input ($16\times16$, bicubic-upsampled for visualization), \textbf{(middle)} super-resolved prediction from the MAE-pretrained model ($64\times64$), and \textbf{(right)} ground-truth high-resolution image ($64\times64$). The MAE-pretrained model preserves fine arc details and structural features more faithfully than the scratch baseline.}
  \label{fig:sr-examples}
\end{figure}

Figure~\ref{fig:sr-examples} shows visual examples of super-resolution outputs.

Figure~\ref{fig:sr-examples} provides qualitative visual comparisons between the low-resolution inputs, super-resolved outputs, and ground-truth high-resolution images. The figure displays multiple example cases from the \texttt{no\_sub} scenario, allowing readers to visually assess the quality of reconstruction across different lensing configurations. Each row presents a side-by-side comparison that highlights both the successes and limitations of the super-resolution model.

Visual inspection reveals that the MAE-pretrained model successfully recovers the overall structure of the Einstein rings and the general brightness distribution. The characteristic arc patterns of strong lensing are clearly preserved in the super-resolved images, demonstrating that the model has learned the fundamental geometric properties of gravitational lensing. The smooth gradients in surface brightness are also well reproduced, indicating accurate reconstruction of the extended light distribution.

However, close examination also reveals areas for potential improvement. Fine substructure features, which carry important information about dark matter, are challenging to recover perfectly from the heavily downsampled inputs. Some high-frequency details in the ground-truth images appear slightly smoother in the reconstructions. These observations motivate future work on incorporating physics-informed priors or adversarial training objectives to better preserve the small-scale features most relevant for dark matter characterization.

\subsection{Mask Ratio Ablation: Improving MAE Beyond Scratch}
\label{sec:ablation}

The mask ratio during MAE pretraining is a critical hyperparameter that controls the difficulty of the self-supervised reconstruction task. This section investigates how varying the mask ratio affects both downstream classification and super-resolution performance. Through systematic ablation experiments, we demonstrate that appropriate mask ratio selection can enable MAE-pretrained models to surpass scratch-trained baselines.

Table~\ref{tab:mask-ablation} reports the effect of different mask ratios during MAE pretraining on both downstream tasks. All metrics in this table are for models using MAE-pretrained encoders followed by task-specific fine-tuning.

\begin{table}[H] 
\caption{Ablation on MAE mask ratio using the same optimizer settings as in Section~\ref{sec:dataset}, with 5 epochs of fine-tuning for each configuration. All rows use MAE-pretrained encoders with task-specific fine-tuning. For the 75\% mask baseline results, see Table~\ref{tab:cls-results}. At 90\% mask, classification AUC (0.9681) and accuracy (88.65\%) exceed the scratch baseline (AUC 0.9567, accuracy 82.46\%).}
\label{tab:mask-ablation}
\centering
\resizebox{\textwidth}{!}{
\begin{tabular}{ccccccc}
\toprule
Mask Ratio & MAE Loss & CLS AUC & CLS Acc & CLS F1 & SR PSNR [dB] & SR SSIM \\
\midrule
50\% & 0.0026 & 0.9502 & 0.7936 & 0.7988 & 34.01 & 0.9544 \\
90\% & 0.0045 & 0.9681 & 0.8865 & 0.8852 & 32.65 & 0.9550 \\
\bottomrule
\end{tabular}
}
\end{table}

Table~\ref{tab:mask-ablation} presents a systematic comparison of MAE pretraining with different mask ratios and their impact on downstream task performance. The table includes metrics for both classification (AUC, accuracy, F1 score) and super-resolution (PSNR, SSIM), as well as the MAE reconstruction loss during pretraining. This comprehensive view reveals important relationships between the pretraining objective and downstream task performance.

The results demonstrate a clear trade-off between classification and super-resolution performance as the mask ratio varies. At lower mask ratios (50\%), the pretraining task is easier, leading to lower reconstruction loss and better super-resolution metrics. Conversely, higher mask ratios (90\%) create a more challenging pretraining objective that forces the encoder to learn more abstract, semantically meaningful features. These high-level features prove more beneficial for the classification task, which requires understanding global image characteristics rather than local pixel details.

The most striking finding is that the 90\% mask ratio configuration achieves classification performance (AUC 0.9681, accuracy 88.65\%) that substantially exceeds the scratch baseline (AUC 0.9567, accuracy 82.46\%). This result validates the hypothesis that aggressive masking during self-supervised pretraining can lead to more transferable representations for discriminative tasks. The encoder, forced to reconstruct images from only 10\% of the patches, must develop a sophisticated understanding of strong-lensing image statistics that proves valuable for downstream classification.

We observe a consistent trade-off between tasks as mask ratio varies:
\begin{itemize}
    \item At \textbf{50\% masking}, the MAE achieves the lowest reconstruction loss (0.0026) and highest SR PSNR (34.01~dB), but classification AUC is 0.9502 with accuracy 79.36\%.
    \item At \textbf{90\% masking}, classification performance is best: AUC 0.9681 and accuracy 88.65\%, exceeding the scratch baseline (AUC 0.9567, accuracy 82.46\%). However, SR PSNR drops to 32.65~dB.
\end{itemize}

These results suggest that more aggressive masking encourages the encoder to learn global, discriminative features beneficial for classification, while lower masking preserves fine details important for reconstruction. The key finding is that \textbf{with appropriate mask ratio tuning (90\%) and the training configuration described in Section~\ref{sec:dataset}, the MAE-pretrained encoder can surpass scratch training on classification} while remaining competitive on super-resolution.

\begin{figure}[H] 
  \centering
  \includegraphics[width=0.65\textwidth]{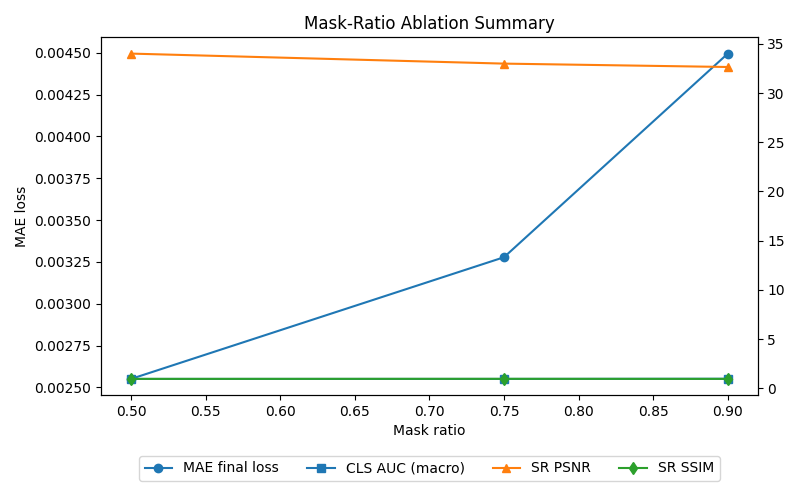}
  \caption{Summary of mask ratio ablation study. This figure visualizes the trade-off between classification metrics (AUC, accuracy) and super-resolution metrics (PSNR, SSIM) as the MAE mask ratio varies from 50\% to 90\%. Higher mask ratios favor classification performance at the cost of slightly degraded reconstruction quality.}
  \label{fig:mask-ablation}
\end{figure}

Figure~\ref{fig:mask-ablation} visualizes these trade-offs.

Figure~\ref{fig:mask-ablation} provides a visual summary of how mask ratio affects performance across both downstream tasks. The plot displays classification and super-resolution metrics on complementary axes, clearly illustrating the trade-off between these two objectives. The visualization enables practitioners to quickly identify the optimal mask ratio for their specific application requirements.

The curves in this figure reveal that the relationship between mask ratio and performance is not monotonic for all metrics. Classification performance shows a clear upward trend with increasing mask ratio, while super-resolution performance decreases. This divergent behavior suggests that the two tasks benefit from fundamentally different types of learned representations, which has important implications for multi-task learning approaches.

From a practical standpoint, this ablation study provides clear guidance for researchers working on similar problems. When the primary goal is classification or detection tasks that require semantic understanding, higher mask ratios (around 90\%) are recommended. When the focus is on image restoration or enhancement tasks, lower mask ratios (around 50\%) better preserve the fine-grained features necessary for high-quality reconstruction. For applications requiring balanced performance on both task types, intermediate mask ratios may offer the best compromise.

\subsection{Feature Space Visualization}

Understanding the structure of learned representations provides valuable insights into model behavior. This section presents t-SNE visualizations of the feature space learned by the MAE-pretrained encoder. These visualizations reveal how well the encoder separates different dark matter classes and help explain the classification performance observed in previous sections.

\begin{figure}[H] 
  \centering
  \includegraphics[width=0.7\textwidth]{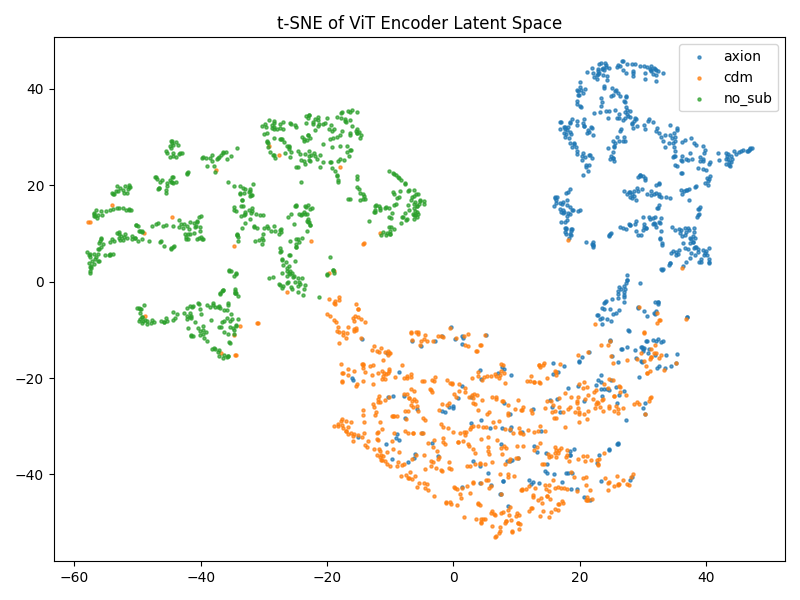}
  \caption{t-SNE visualization of MAE-pretrained encoder features on the test set. Each point represents a single strong-lensing image, colored by its dark-matter class: \texttt{no\_sub} (no subhalos), \texttt{cdm} ($\Lambda$CDM subhalos), or \texttt{axion} (axion-like/wave-like DM). The \texttt{no\_sub} class forms a distinct cluster, while the CDM and axion classes partially overlap, reflecting the subtlety of distinguishing these two substructure scenarios.}
  \label{fig:tsne}
\end{figure}

Figure~\ref{fig:tsne} shows the t-SNE visualization of learned feature representations from the MAE-pretrained encoder.

Figure~\ref{fig:tsne} displays a two-dimensional t-SNE projection of the high-dimensional feature representations extracted by the MAE-pretrained encoder. Each point in the visualization corresponds to a single strong-lensing image from the test set, with color indicating the true dark matter class. The t-SNE algorithm preserves local neighborhood structure from the original high-dimensional space, allowing us to visually assess how well the encoder separates different classes.

The visualization reveals clear structure in the learned feature space. The \texttt{no\_sub} class (smooth lenses without substructure) forms a relatively compact and well-separated cluster, explaining why this class achieves the highest classification accuracy. The encoder has learned distinctive features for smooth lensing configurations that clearly distinguish them from images with substructure. This separation likely reflects differences in the regularity and symmetry of the brightness distributions between smooth and substructured lenses.

The overlap between the \texttt{cdm} and \texttt{axion} clusters provides insight into the classification challenges observed in the confusion matrix. Both classes represent lenses with substructure, differing primarily in the morphology of the perturbations. The partial overlap in feature space indicates that the encoder finds these two classes more similar to each other than either is to the no-substructure class. This observation aligns with physical expectations and suggests that future improvements may require architectural modifications or additional training strategies specifically designed to distinguish between substructure types.

The t-SNE plot shows that the encoder learns partially separable representations. The \texttt{no\_sub} class forms a relatively distinct cluster, while the \texttt{cdm} and \texttt{axion} classes partially overlap, consistent with our classification results where distinguishing between these two substructure scenarios is more challenging.

We now interpret these empirical findings and discuss their implications for MAE-based representations in strong-lensing analysis.

\section{Discussion}
\label{sec:discussion}

The results in Section~\ref{sec:results} demonstrate the feasibility of masked autoencoder pretraining for strong-lensing image analysis. We now discuss several key insights and their implications.

\textbf{MAE vs.\ scratch: the importance of configuration tuning.} The baseline MAE configuration (75\% mask with default hyperparameters) does not automatically outperform scratch training on classification. However, with refined hyperparameters and increasing the mask ratio to 90\%, we observe improved classification performance (AUC 0.968, accuracy 88.65\%) compared to the scratch baseline (AUC 0.957, accuracy 82.46\%). This suggests that more aggressive masking may encourage the encoder to learn higher-level, more discriminative features. We note that these are single-run results, and the observed improvements, while consistent, should be interpreted with appropriate caution.

\textbf{Apparent trade-off between classification and super-resolution.} Higher mask ratios improve classification but slightly degrade super-resolution quality (PSNR drops from 34.01 to 32.65~dB as mask ratio increases from 50\% to 90\%). Within our experimental setup, this reflects a consistent trade-off: aggressive masking encourages learning abstract features useful for classification, while lower masking preserves fine details needed for accurate reconstruction. Practitioners should choose the mask ratio based on their downstream task priorities.

\textbf{Pretraining data choice.} We pretrained on only the \texttt{no\_sub} class, which tests whether representations learned from smooth lenses generalize to substructured scenarios. This choice may explain why the frozen encoder performs poorly, as it never saw substructure during pretraining. Pretraining on a mixture of all three classes is a natural extension that could improve class separation, especially between CDM and axion scenarios, and is left for future work.

\textbf{Reusable encoder for multiple tasks.} A key advantage of our approach is that the pretrained encoder can be reused for multiple downstream tasks with only task-specific fine-tuning. This is particularly valuable in astrophysics, where labeled data may be scarce or expensive to obtain. The same encoder achieves strong performance on both classification and super-resolution, demonstrating the versatility of MAE-learned representations for strong-lensing analysis.

\textbf{Limitations.} Our study has several limitations: (1) We use simulated data from the DeepLense benchmark, and performance on real observational data remains to be tested. (2) The domain shift between simulations and real telescope images is not addressed. (3) Super-resolution is evaluated only on the \texttt{no\_sub} scenario, and may not capture the behavior for lenses with substructure. (4) We do not experiment with joint multi-task training, which could potentially improve both tasks simultaneously.

\section{Conclusion}
\label{sec:conclusion}

Taken together, our findings suggest that MAE-pretrained encoders are a promising building block for multi-task strong-lensing analysis. We presented a shared-encoder framework using masked autoencoder pretraining, enabling both dark-matter model classification and lensed image super-resolution from a single pretrained ViT encoder that is then fine-tuned separately for each task.

Key findings include:
\begin{itemize}
    \item With appropriate mask ratio tuning (90\%) and the training setup described in Section~\ref{sec:dataset}, the MAE-pretrained encoder achieves classification AUC of 0.968 and accuracy of 88.65\%, compared to a ViT trained from scratch (AUC 0.957, accuracy 82.46\%).
    \item For super-resolution, the pretrained model achieves SSIM of 0.961, modestly exceeding scratch training (SSIM 0.955).
    \item Higher mask ratios improve classification but slightly degrade reconstruction quality, revealing a consistent trade-off in learned representations within our experimental setup.
    \item Fine-tuning is essential, as frozen pretrained features are not linearly separable for dark matter classification.
    \item The pretrained encoder learns partially class-separable representations, with clearer separation for the no-substructure class.
\end{itemize}

In summary, MAE pretraining yields a single encoder that can be reused for both dark-matter classification and strong-lensing image super-resolution. With appropriately chosen mask ratios and hyperparameters, this shared encoder can match or modestly exceed the performance of task-specific models trained from scratch in our experiments, while remaining flexible for additional downstream tasks such as subhalo mass regression or domain adaptation to different survey conditions.

Future directions include pretraining on all dark matter classes, incorporating physical constraints into the model, domain adaptation to bridge simulated and real observational data, and applying the framework to upcoming surveys such as Euclid and the Vera C.\ Rubin Observatory Legacy Survey of Space and Time (LSST).

\section*{Acknowledgements}

We thank the DeepLense ML4SCI collaboration for providing the benchmark datasets used in this study. We also thank the developers of \textsc{lenstronomy} for the lens simulation software.

\section*{Funding}

This work received no external funding.

\section*{Declaration of Competing Interests}

The authors declare no competing interests.

\section{Data and Code Availability}
\label{sec:data}

The simulated strong-lensing datasets are available from the DeepLense ML4SCI project at \url{https://ml4sci.org/gsoc/projects/2025/project_DEEPLENSE.html}. Code and trained models will be released at \url{https://github.com/achmadardanip/mae-lensing} upon acceptance.

\bibliographystyle{elsarticle-harv}
\bibliography{BibFile}

\end{document}